\documentclass[lettersize,journal]{IEEEtran}
\usepackage{amsmath,amsfonts}
\usepackage{algorithmic}
\usepackage{algorithm}
\usepackage{array}
\usepackage[caption=false,font=normalsize,labelfont=sf,textfont=sf]{subfig}
\usepackage{textcomp}
\usepackage{stfloats}
\usepackage{url}
\usepackage{verbatim}
\usepackage{graphicx}
\usepackage{cite}
\begin{document}

\title{YOLOv1 to YOLOv10: The fastest and most accurate real-time object detection systems}

\author{Chien-Yao Wang$^{1,2}$ and Hong-Yuan Mark Liao$^{1,2,3}$ \\
        $^{1}$Institute of Information Science, Academia Sinica, Taiwan \\
        $^{2}$National Taipei University of Technology, Taiwan \\
        $^{3}$National Chung Hsing University, Taiwan \\
        \{kinyiu, liao\}@iis.sinica.edu.tw
}

\markboth{YOLO Survey, August~2024}%
{Shell \MakeLowercase{\textit{et al.}}: A Sample Article Using IEEEtran.cls for IEEE Journals}


\maketitle

\begin{abstract}
This is a comprehensive review of the YOLO series of systems. Different from previous literature surveys, this review article re-examines the characteristics of the YOLO series from the latest technical point of view. At the same time, we also analyzed how the YOLO series continued to influence and promote real-time computer vision-related research and led to the subsequent development of computer vision and language models. We take a closer look at how the methods proposed by the YOLO series in the past ten years have affected the development of subsequent technologies and show the applications of YOLO in various fields. We hope this article can play a good guiding role in subsequent real-time computer vision development. 
\end{abstract}

\begin{IEEEkeywords}
YOLO, computer vision, real-time object detection.
\end{IEEEkeywords}

\section{Introduction}
\label{sec:intro}

Object detection is a fundamental computer vision task that can support a wide range of downstream tasks.  For example, it can be used to assist instance segmentation, multi-object tracking, behavior analysis and recognition, face recognition, etc.  Therefore, it has been a popular research topic in the past few decades.  In recent years, due to the popularity of mobile devices, the ability to perform real-time object detection on the edge has become a necessary component for various real-world applications.  Tasks belonging to such applications include autonomous driving, industrial robots, identity authentication, smart health care, visual surveillance, etc.  Among the many real-time object detection algorithms, the YOLO (You Only Look Once) series (from v1 to v10)~\cite{redmon2016you,redmon2017yolo9000,redmon2018yolov3,bochkovskiy2020yolov4,glenn2022yolov5,li2022yolov6,wang2023yolov7,glenn2024yolov8,wang2024yolov9,wang2024yolov10} developed in recent years is particularly outstanding.  It has greatly and extensively affected various research in the field of computer vision.  This paper will review the YOLO family of technologies and their impact on the development of contemporary real-time computer vision systems.

The first deep learning-based method to achieve breakthrough success in the field of object detection was R-CNN~\cite{girshick2014rich}.  R-CNN is a two-stage object detection method, which divides the object detection process into two stages: object proposal generation and object proposal classification.  What R-CNN does is to first use selective search~\cite{uijlings2013selective}, which is commonly used in image processing, to extract proposals.  At this stage, CNN is only used as a feature extractor to extract features of proposals.  As for the recognition part, SVM~\cite{hearst1998support} is used.  The subsequent development of Fast R-CNN~\cite{girshick2015fast} and Faster R-CNN~\cite{ren2015faster} respectively used SPPNet~\cite{he2015spatial} to accelerate feature extraction and proposed Region Proposal Networks to gradually convert object detection into the end-to-end format.  YOLO~\cite{redmon2016you} was proposed by Joseph Radmon in 2015.  It uses per gird prediction to complete object detection in one step.  This is a groundbreaking approach that brings the field of real-time object detection to a whole new level.  The subsequent development of classic one-stage object detection systems includes SSD~\cite{liu2016ssd}, RetinaNet~\cite{lin2017focal}, FCOS~\cite{tian2020fcos}, etc.

Although the one-stage object detection method can detect objects in real time, there is still a gap in accuracy from the two-stage object detection method.  The one-stage detection systems such as RetinaNet~\cite{lin2017focal} and YOLOv3~\cite{redmon2018yolov3} have made significant progress on this issue, and they both achieved sufficient accuracy. YOLO series have become the most preferred method for industry and all academia and research centers that require real-time object analysis.  In 2020, scaled-YOLOv4~\cite{wang2021scaled} further designed a very effective object detection model scaling method.  For the first time, the accuracy of the one-stage object detection method in the field of general object detection surpassed all contemporary two-stage object detection methods, and this achievement also led to many subsequent related research based on YOLO series methods.

In addition to object detection, YOLO series is also used in other computer vision fields as a basis for developing real-time systems.  Currently in instance segmentation, pose estimation, image segmentation, 3D object detection, open-vocabulary object detection, etc., YOLO still plays a pivotal role in real-time systems.

In this review article, we will introduce the following issues in order:

\begin{itemize}
    \item Introduction to the YOLO series methods and important related literature.
    \item The impact of the YOLO family of methods on the contemporary field of computer vision.
    \item Important methods for applying YOLO in different computer vision fields.
\end{itemize}

\section{YOLO series}
\label{sec:yolo}

YOLO is synonymous with the most advanced real-time object detector of our time.  The biggest difference between YOLO and traditional object detection systems is that it abandons the previous two-stage object detection method that requires first finding the locations where objects may be located in the image, and then analyzing the content of these locations individually.  YOLO proposes a unified one-stage object detection method, and this method is streamlined and efficient, which makes YOLO widely used in various edge devices and real-time applications.  Next we will introduce several representative YOLO versions, and this literature review is different from the previous ones.  We will put our emphasis on the state-of-the-art object detection methods and review the advantages and disadvantages of these methods.

\begin{figure}[h]
	\begin{center}
		\includegraphics[width=1.\linewidth]{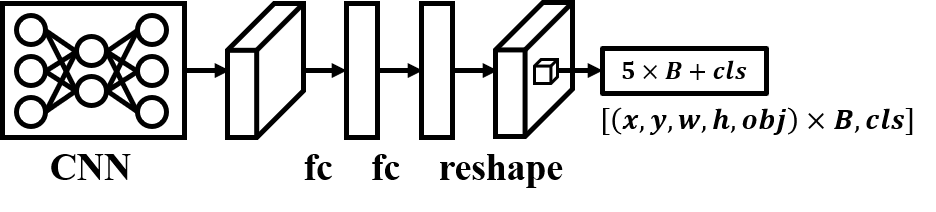}
	\end{center}
    \vspace{-16pt}
	\caption{Architecture of YOLOv1.}		
	\label{fig:yolov1}
    \vspace{-16pt}
\end{figure}

\subsection{YOLO (YOLOv1)} 
\label{sec:yolov1}

YOLOv1: Redmon et al. ~\cite{redmon2016you} was the first one who proposed the one-stage object detector in 2015, and the architecture of YOLOv1 is illustrated in Figure~\ref{fig:yolov1}.  As shown in the figure, an input image first passes through CNN for feature extraction, and then passes through two fully connected layers to obtain global features.  Then, the aforementioned global features are reshaped back to the two-dimensional space for per grid prediction.  YOLOv1 has the following important features:

\noindent{\textbf{One-Stage Object Detector.}} As shown in Figure~\ref{fig:yolov1}, YOLOv1 directly classifies each grid of feature map, and also predicts $B$ bounding boxes.  Each bounding box will predict the object center $(b_x, b_y)$, object size $(b_w, b_h)$, and object score $(b_{obj})$ respectively.  The one stage prediction method does not need to rely on the selective search that must be executed in the object proposal generation stage, which can avoid missed detections caused by insufficient manual design clues.  In addition, the one-stage method can avoid the large number of parameters and calculations generated by fully connected layers in the second stage, and it can avoid the irregular operations required when connecting two stages of RoI operations.  Therefore, YOLO's design can capture features and make predictions more timely and effectively.  Below we will take a closer look at the most important concepts in YOLOv1, which are anchor-free bounding box regression, IoU-aware objectness, and global context features.

\noindent{\textbf{Anchor-free Bounding Box Regression.}} In Equation~\ref{eq:v1}, YOLOv1 directly predicts the proportion of the length and width of the object in the entire image.  Although the anchor-free method requires optimization of a large dynamic range of length and width, which makes convergence more difficult, it also has the advantage of being able to predict some special examples more accurately because it is not restricted by anchors.

\begin{equation}
    \begin{aligned}
    & b_x = t_x+c_x, \\
    & b_y = t_y+c_y, \\
    & b_w = {t_w}^2, \\
    & b_h = {t_h}^2
    \end{aligned}
    \label{eq:v1}
\end{equation}

\noindent{\textbf{IoU-aware Objectness.}} In order to more accurately measure the quality of bounding box prediction, the method proposed by YOLOv1 is to predict the IoU value between a certain bounding  box and the assigned ground-truth bounding box, and use this as the soft label of the objectness predicted by IoU-aware branch.  Finally, the confidence score of bounding box is determined by the product of objectness score and classification probability.

\noindent{\textbf{Global Context Feature.}} To ensure that a grid doesn't only see the local feature and cause prediction errors, YOLOv1 uses fully connected layer to retrieve global context features.  In such a design, no matter what the underlying CNN architecture is, each grid can see a sufficient range of features to predict the target object during prediction.  Compared with fast R-CNN, this design effectively reduces background error by more than half.

\begin{figure}[h]
	\begin{center}
		\includegraphics[width=1.\linewidth]{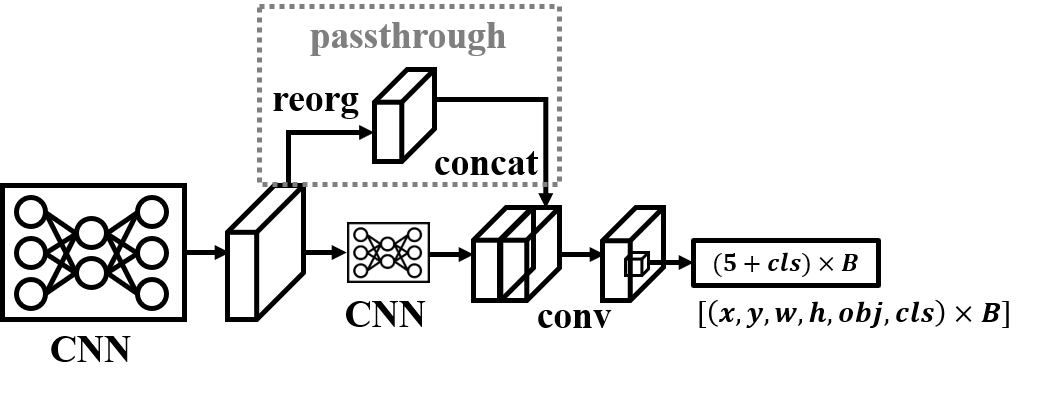}
	\end{center}
    \vspace{-16pt}
	\caption{Architecture of YOLOv2.}		
	\label{fig:yolov2}
    \vspace{-16pt}
\end{figure}

\subsection{YOLO9000 (YOLOv2)} 
\label{sec:yolov2}

In addition to proposing many insightful new methods, Redmon and Farhadi~\cite{redmon2017yolo9000} also integrate various existing techniques.  They designed an object detector that combines high accuracy and speed, as shown in Figure~\ref{fig:yolov2}.  They converted the entire object detection architecture to full convolutional network.  They then combined high-resolution and low-resolution features, and finally use anchor-based for prediction.  Due to its simple input and output formats, YOLOv2 is still one of the mainstream object detection methods commonly used in maintenance and development of many industrial scenes, especially on low-end devices with very limited computing resources.  Below we will discuss the most essential parts of YOLOv2 respectively regarding dimension cluster, direct location prediction, fine-grained feature, resolution calibration, and joint training with WordTree issues.

\noindent{\textbf{Dimension Cluster.}} YOLOv2 proposed to use IoU distance as the basis to k-means clustering on ground-truth bounding boxes to obtain anchors.  On the one hand, the anchor obtained by using dimension cluster can avoid the original manually set aspect ratio, which is difficult to learn the bounding box prediction of the object.  On the other hand, it is also easier to converge than the bounding box regression of the anchor-free approach.

\noindent{\textbf{Direct Location Prediction.}} Faster R-CNN uses the anchor center as the basis to predict the offset between the object center and the anchor center.  The above approach is very unstable during early training.  YOLOv2 follows the object center regression method of YOLOv1 and directly predicts the true position of the object center based on the upper left corner of the grid responsible for predicting the object.

\noindent{\textbf{Fine-grained Feature.}} Passthrough layer predicts by reorganizing high-resolution features into lossless spatial-to-depth and combining them with low-resolution features.  This can enhance fine-grained small object detection capabilities through high-resolution features while taking into account speed simultaneously.

\noindent{\textbf{Resolution Calibration.}} Since the backbone CNN often uses lower-resolution images for image classification pre-training than those used for object detection training, the pre-trained model has never seen the state of larger objects. YOLOv2 uses the image classification pre-train of the same training size image, so that the object detection training process does not require additional learning of new size object information.

\noindent{\textbf{Joint Training with WordTree.}} YOLOv2 designed the training of group softmax using ImageNet with a similar hierarchy as WordTree, and then integrated the categories of COCO and ImageNet using WordTree.  In the end, this technology requires joint training of ImageNet's image classification and COCO's object detection tasks.  Because of the above design, YOLOv2 has the ability to detect 9000 categories of objects.

\begin{figure}[h]
	\begin{center}
		\includegraphics[width=1.\linewidth]{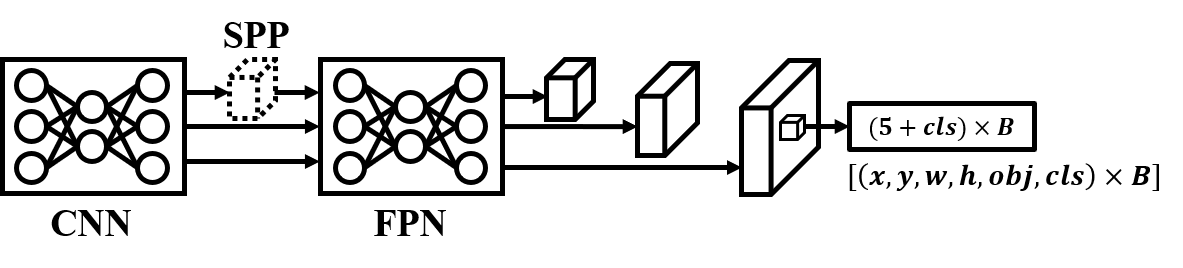}
	\end{center}
    \vspace{-16pt}
	\caption{Architecture of YOLOv3, YOLOv5, and PP-YOLO.}		
	\label{fig:yolov3}
    \vspace{-16pt}
\end{figure}

\subsection{YOLOv3} 
\label{sec:yolov3}

YOLOv3~\cite{redmon2018yolov3} was proposed by Redmon and Farhadi in 2018.  They integrated the advanced technology of existing object detection and made corresponding optimizations to one-stage object detectors.  As shown in Figure 3, in terms of architecture, YOLOv3 mainly combines FPN~\cite{lin2017feature} to enable prediction of multiple scales at the same time.  It also introduces the residual network architecture and designs DarkNet53.  In addition, YOLOv3 also made significant changes to the label assignment task.  The first change is that a ground truth will only be assigned to one anchor, while the second change is to change from soft label to hard label for IoU aware objectness.  To this day YOLOv3 is still the most popular version of YOLO series.  In what follows, let us detail the special designs of YOLOv3, namely prediction across scales, high GPU utility, and SPP.

\noindent{\textbf{Predictions Across Scales.}} YOLOv3 combines FPN to achieve prediction across scales, which can greatly improve the detection ability of small objects.

\noindent{\textbf{High GPU Utility.}} In the time of 2018, mainstream network architecture design focuses on reducing the amount of calculations and parameters.  The design of Draknet53 has higher GPU hardware utilization than other architectures, so it has faster inference speed under the same amount of calculations.  Such a design has also led to subsequent architectural research focusing on actual hardware inference speed.

\noindent{\textbf{SPP.}} YOLOv1 uses fully connected layer to obtain global context features, while YOLOv2 uses passthrough layer to combine multiple resolution features.  YOLOv3 designed multiple maximum pooling layers with a stride of 1 for kernel size from local to global. This design allows each grid to obtain multiple resolution features from local to global.  SPP has been proven to be a simple, efficient method that can greatly improve accuracy.

\begin{figure}[h]
	\begin{center}
		\includegraphics[width=1.\linewidth]{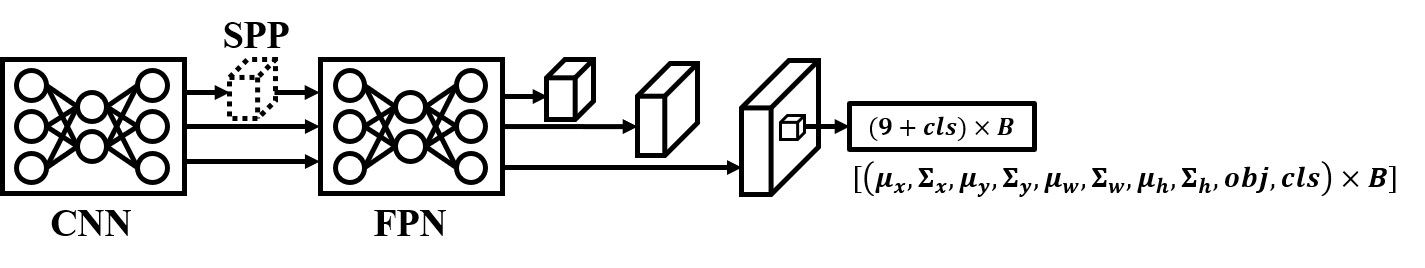}
	\end{center}
    \vspace{-16pt}
	\caption{Architecture of Gaussian YOLOv3.}		
	\label{fig:gyolov3}
    \vspace{-16pt}
\end{figure}

\subsection{Gaussian YOLOv3} 
\label{sec:gyolo}

Gaussian YOLOv3~\cite{choi2019gaussian} proposed a great way to significantly reduce the false positive of an object detection process.  Gaussian YOLOv3 mainly changes the decoding method of the prediction head, and the method used is to convert the bounding box numerical regression problem into predicting its distribution.  Its architecture is shown in Figure~\ref{fig:gyolov3}.  

\noindent{\textbf{Distribution-Based Bounding Box Regression.}} The distribution-based bounding box regression module included in the figure is the uncertainty of predicting bounding box $(x, y, w, h)$ of a Gaussian distribution.  This prediction method can significantly reduce the false positive of object detection.

\begin{figure}[h]
	\begin{center}
		\includegraphics[width=1.\linewidth]{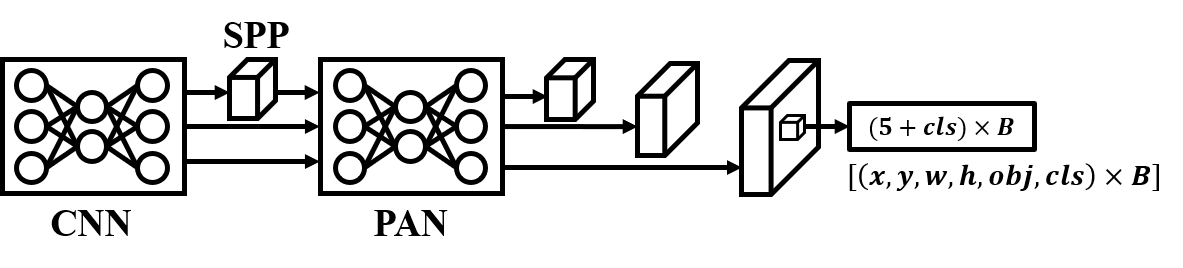}
	\end{center}
    \vspace{-16pt}
	\caption{Architecture of YOLOv4, Scaled-YOLOv4, YOLOv5 r1--r7, and PP-YOLOv2.}		
	\label{fig:yolov4}
    \vspace{-16pt}
\end{figure}

\subsection{YOLOv4} 
\label{sec:yolov4}

Since Joseph Redmon withdrew from computer vision research for some reason, subsequent versions of YOLO were mainly released on the open source platform GitHub.  As for the publication time of the paper, it is later than the open source time.  YOLOv4~\cite{bochkovskiy2020yolov4} was submitted to Joseph Redmon as a draft by Alexey Bochkovskiy, in early April 2020, and was officially released on April 23 2020.  YOLOv4 mainly integrates various technologies in different fields of computer vision in recent years to improve the learning effect of real-time object detectors.  The architectural change of YOLOv4 is to replace FPN with PAN~\cite{liu2018path} and introduce CSPNet~\cite{wang2020cspnet} as backbone.  Most of the subsequent similar YOLO architectures followed this architecture.  In view of the rapid innovation of deep learning technology YOLOv4 was not only developed based on DarkNet~\cite{pjreddie2018darknet,alexeyab2019darknet}, but also implemented on the most successful PyTorch YOLOv3~\cite{glenn2019yolov3} at the time.  YOLOv4 successfully demonstrated how to use one GPU to train an object detector that is as accurate as those trained with more than 128 GPUs, and at the same time has more than three times the inference speed.  The excellent performance of YOLOv4 has also led to many subsequent object detection research.  The following is a list of new features developed in YOLOv4:

\noindent{\textbf{Bag of Freebies.}} YOLOv4 introduces training techniques that only increase training time but do not affect inference time mainly including loss function, regularization methods, data augmentation methods, and label assignment methods.

\noindent{\textbf{Bag of Specials.}} YOLOv4 also introduces methods that only slightly affect the inference time but can greatly improve the accuracy, mainly including receptive field modules, attention mechanism, activation function, and normalization layers, and select useful combinations to add to the system.

\noindent{\textbf{Grid Sensitive Decoder.}} Users on the open source platform found that it was difficult to predict accurately when the object center was near the grid boundary.  YOLOv4 analyzed the reason and found that the gradient from Sigmoid function would approach zero at extreme values.  The developers of YOLOv4 then designed a decoding method as shown in Equation~\ref{eq:v4} to make the predicted target values fall within the effective gradient range.

\begin{equation}
    \begin{aligned}
    & b_x = (1+s_x) \sigma(t_x)-0.5 s_x+c_x, \\
    & b_y = (1+s_y) \sigma(t_y)-0.5 s_y+c_y, \\
    & b_w = p_w e^{t_w}, \\
    & b_h = p_h e^{t_h}
    \end{aligned}
    \label{eq:v4}
\end{equation}

\noindent{\textbf{Self-Adversarial Training.}} YOLOv4 also introduces self-adversarial sample generation training to enhance the robustness of the object detection system.

\noindent{\textbf{Training with Memory Sharing.}} YOLOv4 is also designed to allow GPU and CPU to share memory for storing the information required for gradient updates.  This design allows the trained batch size no longer be limited by GPU memory.

\subsection{Scaled-YOLOv4} 
\label{sec:syolov4}

In 2020, Wang et al.~\cite{wang2021scaled} continued the success achieved with YOLOv4 and continued to develop scaled-YOLOv4 that can be used on both edge and cloud.  Thanks to the activity of the DarkNet and PyTorch YOLOv3 communities, scaled-YOLOv4 can abandon the pre-train steps required by ImageNet and directly use the train-from-scratch method to obtain high-quality object detection results.  In terms of architecture, scaled-YOLOv4 has also introduced CSPNet into PAN, which can comprehensively improve the performance of speed, accuracy, number of parameters, and number of calculations.  Scaled-YOLOv4 also designs model scaling methods for various edge devices and provides three types of models: P5, P6, and P7.  In the training part, scaled-YOLOv4 uses the decoder and label assignment strategy proposed by the initial version of YOLOv5.  Because of the various improvements mentioned above, scaled-YOLOv4 has achieved the highest accuracy and fastest inference speed of all object detectors.  Below we list several unique designs of scaled-YOLOv4:

\noindent{\textbf{Compound Model Scaling.}} Previous model scaling methods only considered the integer hyperparameters of a given architecture.  Scaled-YOLOv4 proposed a model scaling that simultaneously considers the input image resolution and receptive field matching, and uses the number of scaling model stages to design a more efficient architecture that can be applied to high-resolution images.

\noindent{\textbf{Hardware Friendly Architecture.}} Taking into account ShuffleNetv2 \cite{ma2018shufflenet} and HardNet's~\cite{chao2019hardnet} analysis of hardware performance, the highly efficient CSPDark module and CSPOSA module were designed.

\noindent{\textbf{Na\"ive Once For All Model.}} Since scaled-YOLOv4 is trained in the mode of train-from-scratch, the problem of inconsistent resolution between the pre-trained models and the detection model no longer exists.  However, the problem of inconsistency between user input images and training data still exists.  The model scaling method proposed in scaled-YOLOv4 allows users to obtain the best accuracy without re-training during the inference stage, and only needs to remove the output of the corresponding stage.

\subsection{YOLOv5} 
\label{sec:yolov5}

YOLOv5~\cite{glenn2022yolov5} continues the design concept of PyTorch YOLOv3 and has simplified and revised the overall architecture definition method.  So far, there are about 10 different versions.  The initial version is designed with an architecture similar to YOLOv3, while following EfficientDet's~\cite{tan2020efficientdet} model scaling pattern to provide models with different specifications.  PyTorch YOLOv3 was developed from Erik's open source codes~\cite{eriklindernoren2018yolov3}, so it uses GPL3 license, and subsequent versions are adjusted to the more strict AGPL3 license.  YOLOv5 inherits many functions of PyTorch YOLOv3, such as using evolutionary algorithms for auto anchor and hyper-parameter search.  When YOLOv5~\cite{glenn2022yolov5} was open sourced, its performance was slightly worse than YOLOv3-SPP.  After successively combining the CSPNet used by YOLOv4 and the CSPPAN used by scaled-YOLOv4, the first version of YOLOv5 r1.0~\cite{glenn2022yolov5r1} was officially released in June 2020.  Then the developers of YOLOv5 optimized both the speed-accuracy trade-off of the CSP fusion layer and the activation function, quoted YOLOR-based training hyper-parameters, used YOLOv5 r5.0~\cite{glenn2022yolov5r5} in April 2021.  The latest version of YOLOv5 is YOLOv5 r7.0~\cite{glenn2022yolov5r7} Glenn released in November 2022.  Due to the continuous maintenance and version updates by companies, YOLOv5 is currently the most popular YOLO development platform.  Let us point out some distinctive features of YOLOv5 as follows:

\noindent{\textbf{Power-based Decoder.}} The length-width regression system of YOLOv3 uses exponential function to estimate offset, and this approach causes instability during training in some datasets.  YOLOv5 proposed power-based decoder in Equation~\ref{eq:v5} to increase training stability.  Since the output value range of power-based decoder is limited to a certain scaling range of anchor, there will be a theoretically bounded recall. 

\begin{equation}
    \begin{aligned}
    & b_x = 2 \sigma(t_x)-0.5+c_x, \\
    & b_y = 2 \sigma(t_y)-0.5+c_y, \\
    & b_w = p_w (2\sigma(t_w))^2, \\
    & b_h = p_h (2\sigma(t_h))^2
    \end{aligned}
    \label{eq:v5}
\end{equation}

\noindent{\textbf{Neighborhood Positive Samples.}} In order to make up for the deficiency caused by recall, YOLOv5 proposed to add more neighbor grids as positive samples.  At the same time, in order to allow these neighbor grids to correctly predict the center point, they also enlarged the sigmoid scaling coefficient of the YOLOv4 center point decoder.

\begin{figure}[h]
	\begin{center}
		\includegraphics[width=1.\linewidth]{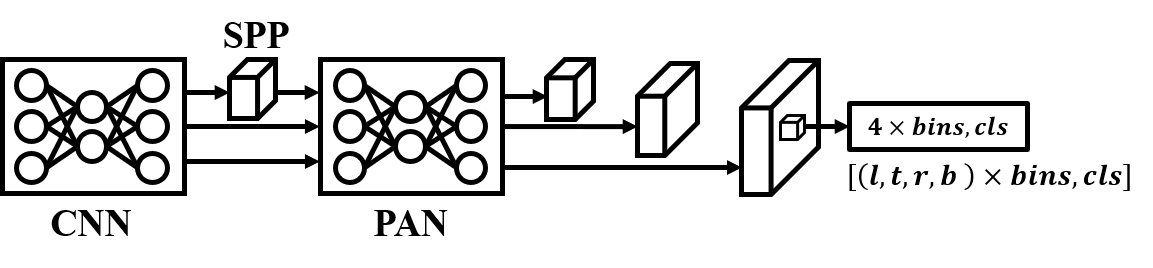}
	\end{center}
    \vspace{-16pt}
	\caption{Architecture of PP-YOLOE, YOLOv6 2.0, YOLOv8, YOLO-NAS.}		
	\label{fig:ppyoloe}
    \vspace{-16pt}
\end{figure}

\newpage

\subsection{PP-YOLO} 
\label{sec:ppyolo}

There are four versions of the PP-YOLO series, namely PP-YOLO~\cite{long2020pp}, PP-YOLOv2~\cite{huang2021pp}, PP-PicoDet~\cite{yu2021pp}, and PP-YOLOE~\cite{xu2022pp}.  PP-YOLO is improved based on YOLOv3.  In addition to using a variety of YOLOv4 training techniques, it also adds CoordConv~\cite{liu2018intriguing}, Matrix NMS~\cite{wang2020solov2}, and better ImageNet pre-trained model and other methods for improvement, while PP-YOLOv2 further introduces scaled-YOLOv4's CSPPAN and other mechanisms.  PP-PicoDet uses neural architecture search as the basis to design the backbone, and introduces YOLOX's anchor-free decoder~\cite{ge2021yolox}.  As for PP-YOLOE, it has made major changes.  It modified RepVGG and designed CSPRepResStage and then used bounding box regression in TOOD's distribution-based regression process~\cite{feng2021tood}.  The YOLO series after YOLOv6 almost all follow the above format.  Listed below are some design features of PP-YOLO series:

\noindent{\textbf{Neural Architecture Search.}} PP-PicoDet is an architecture designed for mobile devices.  It combines ShuffleNetv2~\cite{ma2018shufflenet} and GhostNet~\cite{han2020ghostnet} for one shot neural architecture search.

\noindent{\textbf{Reparameterization Module.}} PP-YOLOE applies RepVGG~\cite{ding2021repvgg} to CSPNet, but removes the identity connection in the training phase.

\noindent{\textbf{Distribution-based Regression Raised.}} PP-YOLOE follows TOOD to use DFL~\cite{li2020generalized} for bounding box regression.  DFL is different from Gaussian YOLOv3 in that it does not need to limit the data to be Gaussian distribution, and can also directly learn the distribution of real data.

\begin{figure}[h]
	\begin{center}
		\includegraphics[width=1.\linewidth]{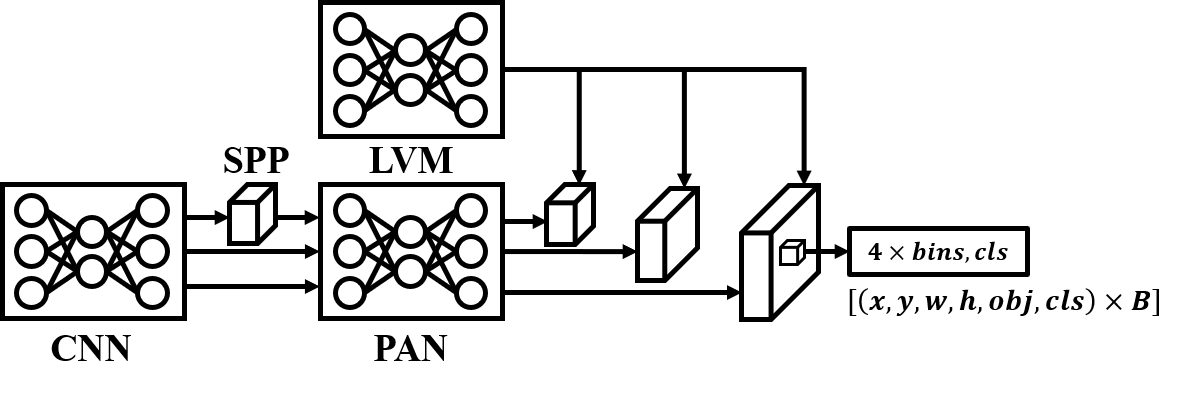}
	\end{center}
    \vspace{-16pt}
	\caption{Architecture of YOLOR.}		
	\label{fig:damoyolo}
    \vspace{-16pt}
\end{figure}

\subsection{YOLOR} 
\label{sec:yolor}

YOLOR~\cite{wang2021you} is not an official version of the YOLO series, but its use of Latent Variable Model (LVM) as implicit knowledge encoder can significantly improve the detection effects of all YOLO series models.  YOLOR's multi-task model has also been widely used in subsequent YOLO versions, and the advanced training technology it proposed has been continued and promoted in all subsequent versions.  Below are some specially designed features of YOLOR:

\noindent{\textbf{Implicit Knowledge Modeling.}} YOLOR proposed three LVMs to encode implicit knowledge, including vector-based, neural network-based, and matrix factorization-based.  The above three encoding methods can effectively enhance the feature alignment, prediction refinement, and multi-task learning capabilities of deep neural networks.

\noindent{\textbf{Multi-task Model.}} YOLOR provides models that can perform object detection, image classification, and multiple object tracking at the same time, and it also provides pose estimation models that based on YOLO-Pose~\cite{maji2022yolo}.

\noindent{\textbf{Advanced Training Technique.}} YOLOR developed advanced autoML technology~\cite{wang2021exploring}, and its techniques for training hyperparameters are continued to be used in the latest version of YOLO series.  YOLOR also uses large dataset pre-train, knowledge distillation, self-supervised learning, and self distillation technologies in its model.  Until now, YOLOR trained using the above method is still the most accurate model of all YOLO series.


\begin{figure}[h]
	\begin{center}
		\includegraphics[width=1.\linewidth]{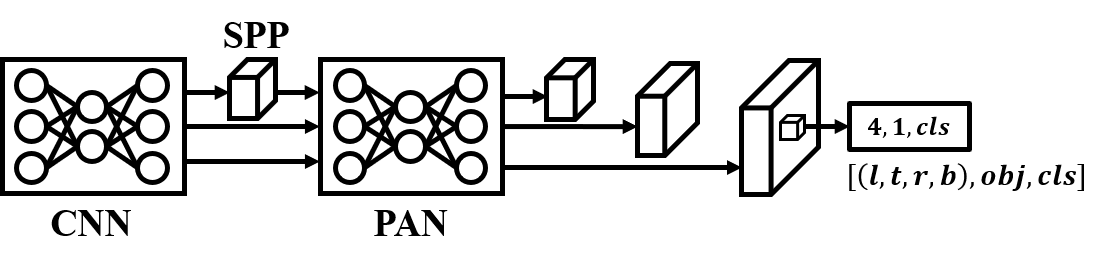}
	\end{center}
    \vspace{-16pt}
	\caption{Architecture of YOLOX.}		
	\label{fig:yolox}
    \vspace{-16pt}
\end{figure}

\subsection{YOLOX} 
\label{sec:yolox}

YOLOX~\cite{ge2021yolox} combined the most practical technologies at the time, mainly based on the CSPNet architecture~\cite{wang2020cspnet} and FCOS' anchor-free head~\cite{tian2020fcos}, improved OTA~\cite{ge2021ota} and proposed the SimOTA dynamic label assignment method to replace the manual label assignment method that was easily confusing.  Subsequent versions of YOLO also began to use or design different dynamic label assignment methods.  The following describes the two most important features of YOLOX:

\noindent{\textbf{Decoupled Head.}} YOLOX uses decoupled head of FCOS, and this design makes classification and bounding box regression easier to learn.

\noindent{\textbf{Anchor-free Strikes Back.}} Due to the development of IoU-based loss, anchor-free head is no longer affected by the loss imbalance caused by the length and width of the object.  With modern technology, anchor-free head can also be well trained.  As for YOLOX, it takes FCOS' anchor-free head to achieve the planned objective.

\begin{figure}[h]
	\begin{center}
		\includegraphics[width=1.\linewidth]{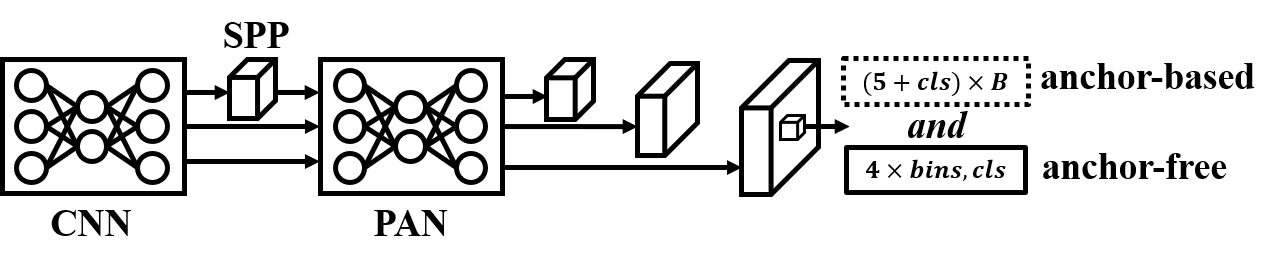}
	\end{center}
    \vspace{-16pt}
	\caption{Architecture of YOLOv6 3.0 and YOLOv6 4.0.}		
	\label{fig:yolov6}
    \vspace{-16pt}
\end{figure}

\subsection{YOLOv6} 
\label{sec:yolov6}

The initial version of YOLOv6~\cite{meituan2022yolov6} uses RepVGG~\cite{ding2021repvgg} as the main architecture.  In versions after version 2.0, such as Li et al. (2022)~\cite{li2022yolov6} and Li et al.(2023)~\cite{li2023yolov6}, CSPNet~\cite{wang2020cspnet} was introduced.  YOLOv6 is a system specially designed for industry, so it has put a lot of effort into quantization issues.  The contributions of YOLOv6 include using RepOPT~\cite{ding2022re} to make the quantized model more stable, and using quantization aware training (QAT) and knowledge distillation to enhance the accuracy of the quantized model.  YOLOv6 version 3.0~\cite{li2023yolov6} proposed a concept of anchor-aid training, as shown in Figure~\ref{fig:yolov6}, to improve the accuracy of the system.  Later in YOLOv6 version 4.0~\cite{li2024yolov}, a lightweight architecture YOLOv6-lite based on depth-wise convolution was proposed to face lower-end computing devices.  The following lists some of the unique features proposed by YOLOv6:

\noindent{\textbf{Reparameterizing Optimizer.}} YOLOv6 version 2.0 uses RepOPT to slow down the accuracy lost after model quantization.

\noindent{\textbf{Quantization Aware Training.}} In YOLOv6 version 2.0, QAT is used to improve the accuracy of the quantization model.

\noindent{\textbf{Knowledge Distillation.}} YOLOv6 version 2.0 uses self-distillation and channel-wise distillation respectively to improve model accuracy, and it also uses QAT to reduce the accuracy loss after model quantization.

\noindent{\textbf{Anchor-Aided Training.}} YOLOv6 version 3.0 proposed using anchor-based head to assist anchor-free head learning, as shown in Figure~\ref{fig:yolov6}, to improve accuracy.

\begin{figure}[h]
	\begin{center}
		\includegraphics[width=1.\linewidth]{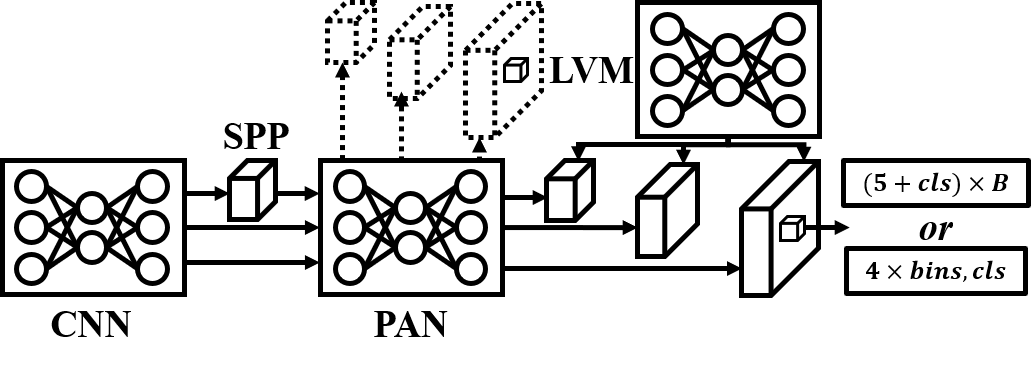}
	\end{center}
    \vspace{-16pt}
	\caption{Architecture of YOLOv7.}		
	\label{fig:yolov7}
    \vspace{-16pt}
\end{figure}

\subsection{YOLOv7} 
\label{sec:yolov7}

YOLOv7~\cite{wang2023yolov7} introduces trainable auxiliary architectures that can be removed or integrated during the inference stage, including YOLOR~\cite{wang2021you}, the recently popular RepVGG~\cite{ding2021repvgg}, and additional auxiliary losses.  Architecturally, YOLOv7 uses ELAN~\cite{wang2022designing} to replace the CSPNet used by YOLOv4, and proposes E-ELAN to design large models.  YOLOv7 also provides a variety of computer vision task-related models and supports anchor-based and anchor-free architectures.  The features of YOLOv7 are listed below:

\noindent{\textbf{Make RepVGG Great Again.}} The reparameterization method proposed by RepVGG~\cite{ding2021repvgg} allows simple network architectures to converge when deepening, but it cannot be effectively applied to modern popular deep network architectures.  The planned RepConv technology proposed by YOLOv7 allows the reparameterization method to effectively bring gains to various residual-based and concatenation-based architectures.

\noindent{\textbf{Consistent Label Assignment.}} The auxiliary loss method used in the past will make the output targets of different branches inconsistent, which will lead to confusion and instability when performing training.  In response to the maturity and popularity of the dynamic label assignment method, YOLOv7 proposed the consistent label assignment mechanism to maintain the consistency of the goals and feature learning directions of different branches.

\noindent{\textbf{Coarse to Fine Label Assignment.}} In the past, multi-stage refinement architectures, such as Cascade R-CNN~\cite{cai2018cascade} and HTC~\cite{chen2019hybrid}, required additional theoretical architectures to refine predictions step by step.  YOLOv7 proposed the coarse-to-fine label assignment mechanism, which can directly use auxiliary loss to guide the coarse-to-fine characteristics in the feature space, providing prediction refinement effects without changing the architecture. 

\noindent{\textbf{Partial Auxiliary Loss.}} YOLOv7 allows some features to receive auxiliary information updates, and the remaining parts are still focused on target task learning.  The developers of YOLOv7 found that this design has a good improvement effect on the main tasks.

\noindent{\textbf{Various Vision Tasks.}} YOLOv7 provides models including object detection, instance segmentation, and pose estimation, and has achieved real-time state-of-the-art in these tasks.

\begin{figure}[h]
	\begin{center}
		\includegraphics[width=1.\linewidth]{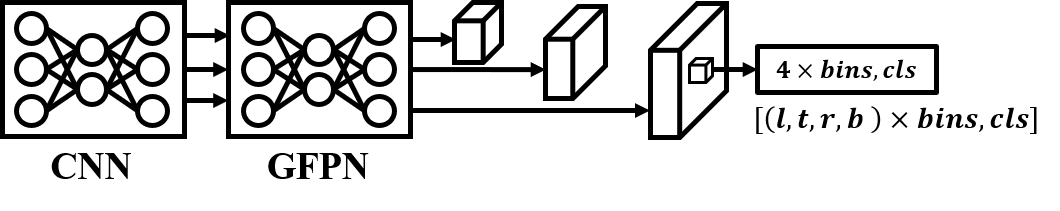}
	\end{center}
    \vspace{-16pt}
	\caption{Architecture of DAMO-YOLO.}		
	\label{fig:damoyolo}
    \vspace{-16pt}
\end{figure}

\subsection{DAMO-YOLO} 
\label{sec:damoyolo}

DAMO-YOLO~\cite{xu2022damo} have proposed improved methods in terms of backbone architecture, feature integration, prediction head, and label assignment.  The block diagram of DAMO-YOLO is shown in Figure~\ref{fig:damoyolo}.  Its features are summarized as follows: 

\noindent{\textbf{MAE-NAS.}} DAMO-YOLO uses MAE-NAS~\cite{sun2021mae} to search CSPNet and ELAN to achieve a more efficient architecture.

\noindent{\textbf{Efficient GFPN.}} DAMO-YOLO disassembled the queen-fusion of GFPN~\cite{jiang2022giraffedet} and retained the fusion layers of the trade-off that is designed for achieving the best speed and accuracy, so as to combine it with ELAN and design RepGFPN.

\noindent{\textbf{ZeroHead.}} DAMO-YOLO simplifies the complex decoupled head into a feature projection layer.

\noindent{\textbf{AlignedOTA.}} DAMO-YOLO proposed aligned OTA to solve the misalignment problem of classification prediction, regression prediction, and dynamic label assignment.

\subsection{YOLOv8} 
\label{sec:yolov8}

YOLOv8~\cite{glenn2024yolov8} is a refactored version of YOLOv5~\cite{glenn2022yolov5}, which updates the way the overall API is used and makes a lot of underlying code optimizations.  It architecturally changes YOLOv7's ELAN, plus additional residual connection, while its decoder is the same as YOLOv6 2.0.  It is not so much a new YOLO version as it is a technology integration platform, and it basically integrates the APIs of multiple downstream tasks and connects them in series.  Its most recent version is Glenn~\cite{glenn2024yolov8}, which integrates the latest technologies such as YOLOv9 and YOLO World~\cite{cheng2024yolo}.  Because program modification and API usage are not very intuitive, many developers have not yet switched to this platform.  But for professional users, the performance improvements brought by optimization of the underlying program code have also attracted many R\&D teams to use it.  In what follows are two special features of YOLOv8:

\noindent{\textbf{Code Optimization.}} The optimization of the underlying program code released by YOLOv8 has brought about 30\% improvement in training performance.

\noindent{\textbf{API for Down-stream Applications.}} YOLOv8 also provides a simple API to connect the detection model with various downstream tasks, such as segment anything, instance segmentation, pose estimation, multiple object tracking, etc.

\subsection{YOLO-NAS} 
\label{sec:yolonas}

YOLO-NAS~\cite{supergradients2023yolonas} did not reveal too many technical details. It mainly uses its own AutoNAC NAS to design the quantization friendly architecture and uses a multi-stage training process, including pre-training on Object365, COCO Pseudo-Labeled data, Knowledge Distillation (KD), and Distribution Focal Loss (DFL).

\begin{figure}[h]
	\begin{center}
		\includegraphics[width=1.\linewidth]{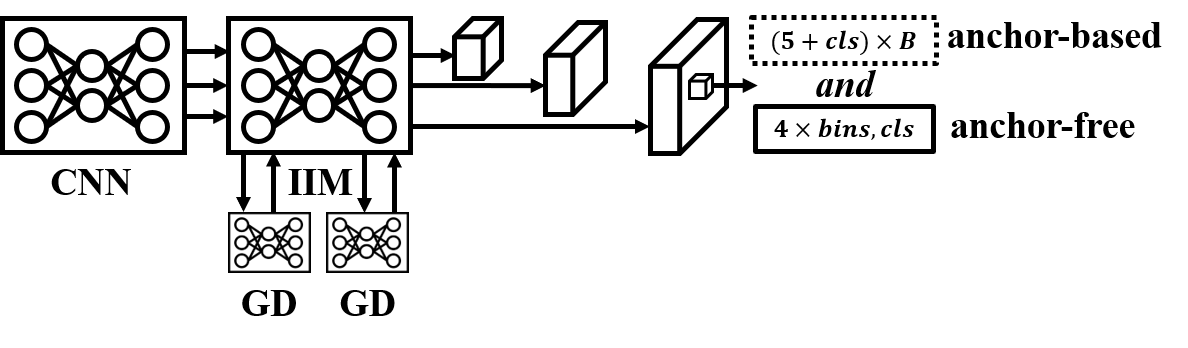}
	\end{center}
    \vspace{-16pt}
	\caption{Architecture of Gold-YOLO.}		
	\label{fig:goldyolo}
    \vspace{-16pt}
\end{figure}

\subsection{Gold-YOLO} 
\label{sec:goldyolo}

Gold-YOLO~\cite{wang2023gold} The overall architecture of Gold-YOLO is similar to that of YOLOv6 3.0. The main design is that the Gather-and-Distribute mechanism replaces PAN in the architecture, and masked image modeling is pre-trained during the training process. 

\noindent{\textbf{Gather-and-Distribute Mechanism.}} The main architecture of Gather-and-Distribute is shown in Figure~\ref{fig:goldyolo}. It mainly collects features from each layer through two gather-and-distribute modules and integrates them into global features using transformers. The integrated global features will be distributed to the low-level and high-level layers respectively, and the distribution method uses the information injection module to integrate the global features with the features that have been distributed to layers. 

\begin{figure}[h]
	\begin{center}
		\includegraphics[width=1.\linewidth]{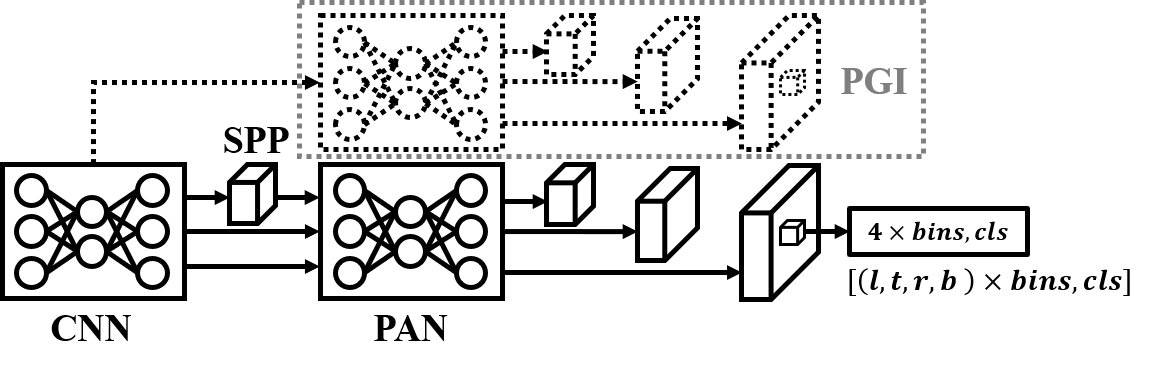}
	\end{center}
    \vspace{-16pt}
	\caption{Architecture of YOLOv9.}		
	\label{fig:yolov9}
    \vspace{-16pt}
\end{figure}

\subsection{YOLOv9} 
\label{sec:yolov9}

YOLOv9~\cite{wang2024yolov9} proposed an important trustworthy technology -- Programmable Gradient Information (PGI), whose architecture is shown in Figure~\ref{fig:yolov9}. The design architecture in the figure can enhance the interpretability, robustness, and versatility of the model. The design of PGI is to use the concepts of reversible architecture and multi-level information to maximize the original data that the model can retain and the information needed to complete the target tasks. YOLOv9 extended ELAN to G-ELAN and used it to show how PGI can achieve excellent accuracy, stability and inference speed on models with low number of parameters. Several outstanding features of YOLOv9 are described below:

\newpage

\noindent{\textbf{Auxiliary Reversible Branch.}} PGI exploits the properties of reversible architecture to solve the information bottleneck problem in deep neural networks. This is completely different from the general-purpose reversible architecture which simply maximizes the information to be retained. What PGI uses is to share the information retained by reversible architecture with the main branch in the form of auxiliary information. On the premise of retaining the information required for the target task, retain as much information as possible from the original data.

\noindent{\textbf{Multi-level Auxiliary Information.}} PGI proposed the concept of multi-level auxiliary information so that each layer of the main branch features retains the information required for all task objectives as much as possible. This can avoid the problem that past methods tend to lose important information at the shallow level, which in turn leads to the inability to obtain sufficient information at the deep level.

\noindent{\textbf{Generalize to Down-stream Tasks.}} Because PGI can maximize the retention of original data information, models trained by PGI achieve more robust performance in small datasets, transfer learning, multi-task learning, and adaptation to new downstream tasks.

\noindent{\textbf{Generalize to Various Architectures.}} PGI can also be applied to other architectures, such as conventional CNN, depth-wise convolutional CNN, transformer, and different types of computer vision methods, such as anchor-based, anchor-free, post-processing free, etc. Therefore, PGI has absolutely superior versatility.

\begin{figure}[h]
	\begin{center}
		\includegraphics[width=1.\linewidth]{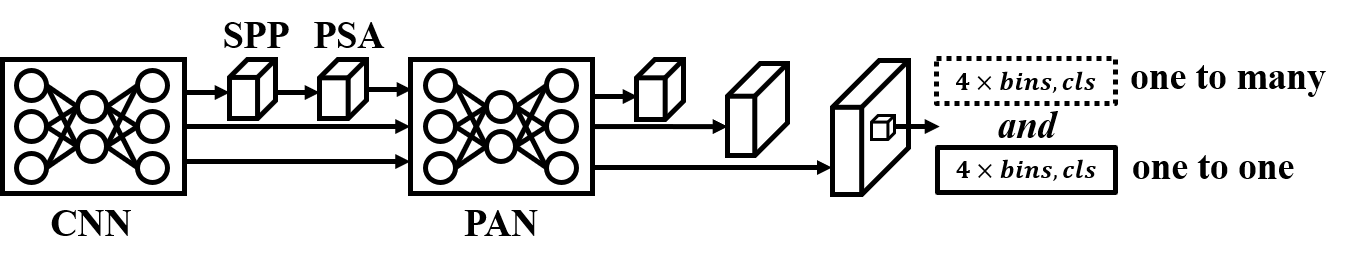}
	\end{center}
    \vspace{-16pt}
	\caption{Architecture of YOLOv10.}		
	\label{fig:yolov10}
    \vspace{-16pt}
\end{figure}

\subsection{YOLOv10} 
\label{sec:yolov10}

The overall architecture of YOLOv10~\cite{wang2024yolov10} is similar to YOLOv6 3.0, but the transformer-based module is added to enhance the extraction of global features. They changed the dual head to one-to-many and one-to-one matching, respectively. This change allows YOLO to do without post-processing likes the DETR-based method, and can directly obtain end-to-end object detection results. Next, we introduce some of the distinct features of YOLOv10.

\noindent{\textbf{Dual Label Assignment.}} Use the label assignment method similar to DATE~\cite{chen2022date}, and add stop gradient operation to the one-to-one branch.

\noindent{\textbf{NMS-free Object Detection.}} The design of one-to-one matching mechanism enables the prediction process without relying on NMS for post-processing.

\noindent{\textbf{Rank-guided Block Design.}} Proposed to use rank to determine which stages use conventional convolution and which stages use depth-wise convolution.

\noindent{\textbf{Partial Self-attention.}} YOLOv10 combined CSPNet and Transformer and proposed the self-attention module.

\newpage

\section{Impact of YOLO series}
\label{sec:impact}

The YOLO series of algorithms have the characteristics of (1) relatively simple frame and (2) relatively easy deployment. In what follows we will describe these characteristics in detail.

\subsection{Simpler}

\noindent{\textbf{Simpler Frame.}} Based on the most forward-looking research on DeepMultiBox~\cite{erhan2014scalable} and OverFeat~\cite{sermanet2014overfeat}, YOLO proposed a new way of one-stage object detection, and this new approach influenced many subsequent computer vision research. Before YOLO series was proposed, the tasks that originally required deep learning to perform dense prediction mainly included pixel-level tasks such as semantic segmentation and optical flow estimation. As for object detection, pose estimation and other instance-level tasks, most of them are split into multiple sub-tasks and predicted in the cascade way. After YOLO was proposed, many algorithms that originally needed to use multi-stage and bottom-up methods were suddenly converted to end-to-end, top-down, and one-stage methods. Examples of this sort include pose estimation and facial landmark detection that directly predict the bounding box and the relative positions of anchor points in the bounding box, as well as multi-object tracking that simultaneously detects and extracts re-identification features.

\noindent{\textbf{Simpler Deployment.}} YOLO does not use specially structured modules, so it is very easy to be deployed on a variety of computing devices. However, there are many special designs such as receptive field module and attention mechanism etc., which are very helpful in improving the accuracy of object detection, but it is not easy to design them into universal and simple modules. YOLO converts these special modules into modules with simple structures through clever design. For example, YOLOv3 proposed to use max-pooling with multiple resolutions to sweep the feature map with a stride of 1 to improve the SPP layer that is originally limited to a fixed input size, and the ASPP layer that requires dilated convolution. The above approach can greatly enhance the model's multi-resolution and global perception capabilities. As for YOLOv4, it proposed to use a convolution layer to replace the small network containing various pooling layers and fully connected layers in the attention module. The entire model of YOLOv4-tiny, YOLOv6 and YOLOv7 was even improved to the point where it only needs to be composed using 1$\times$1 convolution, 3$\times$3 convolution, and max-pooling. In addition, Darknet developed in C language by Joseph Redmon allows the training and inference process of YOLO without relying on additional software packages. The above-mentioned friendly deployment to existing equipment makes the YOLO series widely used in various practical systems.

\subsection{Better}

In addition to being lightweight and easy-to-use as described in the previous section, the YOLO series models also have some more advanced functions, such as better training techniques, better model scalability, and better model generalizability. Below we describe these functions in detail.

\noindent{\textbf{Better Training Technique.}} The training technology proposed by YOLO series models is not only more advanced but also complementary to the most advanced training technology currently available. Many studies in the past have mostly verified the proposed method on a foundation method, such as ResNet~\cite{he2016deep} and ViT~\cite{dosovitskiy2020image} for image recognition, or faster R-CNN~\cite{ren2015faster} and DETR~\cite{carion2020end} for object detection. However, most studies ignore whether the proposed methods are compatible with the current state-of-the-art methods and complement each other to promote the overall progress of the field. Since YOLOv2, the YOLO series has considered compatibility with the most advanced technologies when designing, and at the same time proposed new methods that can complement these technologies. In YOLOv3, YOLOv4, and PP-YOLO series of models, developers also try to analyze technologies that cannot be compatible with each other. This kind of attitude has an important guiding role for subsequent developers.

\noindent{\textbf{Better Model Scalability.}} The YOLO series models do not require special settings when performing model scaling. Scaled-YOLOv4 and YOLOv7 proposed some guidelines for model scaling, while YOLOv5 follows EfficientDet's model scaling method. These scaling methods are directly integrated into the framework, and this kind of design allows users to obtain stable and great performance no matter how they adjust the hyperparameters of model scaling.

\noindent{\textbf{Better Model Generalizability.}} The methods proposed by the YOLO series can be applied to many fields. For example, the concept of using prediction and ground truth to calculate metric proposed by YOLOv1 has been widely used in various soft label generation methods, while the method of using K-means as the initial anchor proposed by YOLOv2 has been extended to the pose estimation field. The WorldTree group softmax method proposed by YOLO is transformed into dealing with the imbalanced data distribution problem of long-tailed learning. SimOTA proposed by YOLOX is used as the basic method for various dynamic label assignment, and the hybrid label assignment method proposed by YOLOv7 is also widely used. The methods proposed by YOLO series are also applicable to a variety of architectures, for example, the CSPNet used by YOLOv4 not only shows excellent results on CNN, but has also been proven to work well with architectures such as Transformer~\cite{vaswani2017attention}, Graph Neural Networks~\cite{scarselli2008graph}, Spiking Neural Networks~\cite{tavanaei2019deep}, and MAMBA~\cite{gu2023mamba}.  The subsequent ELAN used by YOLOv7 has also been rapidly applied in various computer vision fields. 

\subsection{Faster}

\noindent{\textbf{Faster Architecture.}} Another feature of the YOLO series is its very fast inference speed, mainly because its architecture is designed for the actual inference speed of the hardware. The designers of YOLOv3 found that even a simple 1$\times$1 convolution and 3$\times$3 convolution combined architecture, although it has a lower computational load, does not necessarily mean that it has an advantage in inference speed. Therefore, they designed DarkNet for real-time object detection. As for the designers of Scaled-YOLOv4, they referred to research including ShuffleNetv2~\cite{ma2018shufflenet} and HarDNet~\cite{chao2019hardnet}, and further analyzed the criteria that need to be considered for high inference speed architecture for different levels of devices from edge to cloud. To achieve the same purpose, the developers of scaled-YOLOv4 designed Fully CSPOSANet and CSPDarkNet. As for the developers of YOLOv6, they used the efficient RepVGG as the backbone, while the designers of DAMO-YOLO used NAS technology to directly search for efficient architectures in CSPNet and ELAN.

\subsection{Stronger}

\noindent{\textbf{Stronger Adaptability.}} The YOLO series has gained great progress and response in the open source community. The training method integrated by Darknet and PyTorch YOLOv3 allows YOLO series to train object detectors without relying on ImageNet's pre-trained model. Due to the above reasons, the YOLO series can be easily applied to data in different domains without relying on a large number of training models corresponding to the domain. The above advantages enable the YOLO series to be widely used in various application domains. In addition, the YOLO series can also be easily applied to different datasets. For example, PyTorch YOLOv3 proposes to use evolutionary algorithms to automatically search for hyperparameters, which can be applied to different datasets. In addition, the improved anchor-free-based YOLO from YOLOX to PP-YOLOE allows YOLO to rely on fewer hyperparameters during training and can be used more widely in various application domains.

\noindent{\textbf{Stronger Capability.}} The YOLO series has excellent performance in a variety of computer vision tasks. For example, after being widely used in the field of real-time object detection, many other computer vision models based on YOLO have been developed, including YOLACT~\cite{bolya2019yolact} instance segmentation model, JDE~\cite{wang2020towards} multiple object tracking, and so on. Taking YOLOR as an example, it began to combine multiple tasks into the same model for prediction. It can perform image recognition, object detection, and multi-object tracking at the same time, and significantly improve the effect of multi-task joint learning. On the same task, YOLOv5 trains image recognition and object detection models separately. In addition, YOLOv7 also demonstrated outstanding performance in a variety of computer vision domains. At that time, it became the most advanced method for real-time object detection, instance segmentation, and pose estimation. On the same issue, YOLOv8 additionally integrates tasks such as rotating object detection and pose estimation. In addition, YOLOv9 further combines YOLOv7 and YOLOR to extend the multi-task model to the visual-language domain.

\noindent{\textbf{Stronger Versatility.}} Since object detection is a necessary starting step for many practical applications, and as a top object detection method, YOLO's design is very suitable for matching with various downstream task models. In this regard, the design of PP-YOLO series is particularly outstanding, and this series can provide an integrated system for dozens of downstream tasks including face analysis, license plate recognition, multi-object tracking, traffic statistics, behavior analysis, etc.


\section{YOLO for various computer vision tasks}
\label{sec:task}

The YOLO series systems have been widely used in many fields. In this section, we will introduce YOLO's representative works in other computer vision fields and explain the new designs either in architecture or methods completed by these representative works in order to achieve real-time performance. 

\subsection{Multiple Object Tracking}

\noindent{\textbf{ROLO~\cite{guanghan2016rolo}, JDE~\cite{wang2020towards}, CSTrack~\cite{liang2022rethinking}.}}

In the past, deep learning-based multiple object tracking related algorithms, such as Deep-SORT~\cite{wojke2017simple}, needed to crop the detected object area from the original image after detecting objects, and then capture features through additional networks for tracking. ROLO~\cite{guanghan2016rolo} proposes objects directly detected by YOLO, and uses LSTM~\cite{hochreiter1997long} for single object tracking. They proposed to use multiple LSTM to design MOLO and then perform multiple object tracking.  JDE~\cite{wang2020towards} proposes to output the re-ID features for object tracking while detecting objects. However, JDE's multi-scale dense prediction re-ID feature requires a large amount of calculations. In addition, a set of re-ID features will match multiple anchors, making it easy to confuse IDs. CSTrack~\cite{liang2022rethinking} further combines JDE and FairMOT~\cite{zhang2021fairmot}, and after integrating multi-scale features, only outputs re-ID features at one scale. This can achieve more accurate multi-object tracking effects.

\subsection{Instance Segmentation}

\noindent{\textbf{YOLACT~\cite{bolya2019yolact}, YOLACT-Edge~\cite{liu2021yolactedge}, YOLACT++~\cite{zhou2020yolact++}, Insta-YOLO~\cite{mohamed2021insta}, Poly YOLO~\cite{hurtik2022poly}.}}

In the past, most instance segmentation prediction was performed separately for each detected object, so more complex segmentation network is required. YOLACT~\cite{bolya2019yolact} and YOLACT++~\cite{zhou2020yolact++} decompose the instance segmentation process into two steps, namely prototypes and coefficients, and only need to predict coefficients to use these prototypes to form the output instance segmentation results. Using the above method can greatly reduce the amount of operations required when instance segmentation is executed. YOLACTEdge~\cite{liu2021yolactedge} then pushes instance segmentation further to the video domain. The concept of using FeatFlowNet greatly reduces the number of features extracted by the backbone.

Another way to reduce the computation of instance segmentation prediction is to express binary mask in other ways, such as expressing mask in the form of polygon or polar coordinates. Although this expression method will cause some distortion, it can express the mask of the object in very few dimensions. Insta-YOLO~\cite{mohamed2021insta} and Poly YOLO~\cite{hurtik2022poly} are two examples to use the polygon form to predict the result of instance segmentation.

\newpage

\subsection{Automated Driving}

\noindent{\textbf{YOLOP~\cite{wu2022yolop}, YOLOPv2~\cite{han2022yolopv2}, YOLOPv3~\cite{zhan2024multi}, HybridNets~\cite{vu2022hybridnets}, YOLOPX~\cite{zhan2024yolopx}.}}

YOLO series is also widely used in visual perception tasks in self-driving scenarios. YOLOP~\cite{wu2022yolop} and YOLOPv2~\cite{han2022yolopv2} respectively use CSPNet and ELAN as the main architecture for object detection, and therefore can be used for area detection and lane prediction. HybridNet~\cite{vu2022hybridnets}, YOLOPv3~\cite{zhan2024multi}, and YOLOPX~\cite{zhan2024yolopx} are also modified by different versions of YOLO and perform self-driving tasks.

\subsection{Human Pose Estimation}

\noindent{\textbf{KAPAO~\cite{mcnally2022rethinking}, YOLO-Pose~\cite{maji2022yolo}.}}

Human pose estimation can be viewed as additional spatial attributes for predicting object detection targets. Since keypoints do not necessarily fall in grids, additional decoder design is required. KAPAO~\cite{mcnally2022rethinking} divides human pose estimation into human pose object and keypoint object expressions for prediction and combination. YOLO-Pose~\cite{maji2022yolo} directly predicts the regression value of the key relative to the center of the grid, and then execute human pose estimation. The above design can achieve pretty good results.

\subsection{3D Object Detection}

\noindent{\textbf{Complex YOLO~\cite{simony2018complex}, Expandable YOLO~\cite{takahashi2020expandable}, YOLO 6D~\cite{tekin2018real}, YOLO3D~\cite{ruhyadi2022yolo3d}.}}

There are also some studies that generalize YOLO series from 2D to 3D. In addition to ComplexYOLO~\cite{simony2018complex} which combines images and LIDAR as input, and Expandable YOLO~\cite{takahashi2020expandable} which uses RGB-D images as input, there is also YOLO 6D~\cite{tekin2018real} and YOLO 3D~\cite{ruhyadi2022yolo3d} which simply use images as input.

\subsection{Video Perception}

\noindent{\textbf{YOLOV~\cite{shi2023yolov}, YOLOV++~\cite{shi2024practical}, Stream YOLO~\cite{yang2022real}.}}

YOLO series, which performs extremely well in real-time object detection in images, will naturally be applied to the video domain. Among them, YOLOV~\cite{shi2023yolov} and YOLOV++~\cite{shi2024practical} can be applied to video object detection. Alternatively, stream YOLO~\cite{yang2022real} can be used with streaming perception.

\subsection{Face Detection}

\noindent{\textbf{YOLO-Face~\cite{chen2021yolo}, YOLO-Face v2~\cite{yu2024yolo}, YOLO5Face~\cite{qi2022yolo5face}.}}

Face detection is one of the most popular subfields among the various possible application domains of object detection. The face detection models designed based on YOLO also performs quite well in this field.

\subsection{Image Segmentation}

\noindent{\textbf{Fast-SAM~\cite{zhao2023fast}.}}

Due to the real-time and high-performance characteristics of YOLO, it has also begun to be combined with many foundation models and applied to new computer vision tasks. Fast-SAM~\cite{zhao2023fast} combines YOLO with SAM~\cite{kirillov2023segment} and applies it to the general image segmentation task. The above combination can greatly improve the inference speed of the task model.

\subsection{Open Vocabulary Object Detection}

\noindent{\textbf{YOLO-World~\cite{cheng2024yolo}, Open-YOLO 3D~\cite{boudjoghra2024open}.}}

YOLO is also used in conjunction with visual language foundation models.  Examples of this sort include YOLO-world~\cite{cheng2024yolo} and Open-YOLO 3D~\cite{boudjoghra2024open}, which combine YOLO and CLIP~\cite{radford2021learning} methods and can be used to perform 2D and 3D open vocabulary object detection respectively.

\subsection{Combine With Other Architecture}

\noindent{\textbf{ViT-YOLO~\cite{zhang2021vit}, DEYO~\cite{ouyang2022deyo}, DEYOv2~\cite{ouyang2023deyov2}, DEYOv3~\cite{ouyang2023deyov3,ouyang2024deyo}, Mamba-YOLO~\cite{wang2024mamba}, Spiking YOLO~\cite{kim2020spiking}, GNN-YOLO~\cite{gong2024yolo}, GCN-YOLO~\cite{chen2024gcn}, KAN-YOLO~\cite{danielsyahputra2024kanyolo}.}}

YOLO also demonstrates compatibility with a variety of deep neural network architectures.  Architectures of this sort include ViT~\cite{zhang2021vit,ouyang2022deyo,ouyang2023deyov2,ouyang2023deyov3,ouyang2024deyo}, MAMBA~\cite{wang2024mamba}, SNN~\cite{kim2020spiking}, GNN~\cite{gong2024yolo,chen2024gcn}, and KAN~\cite{danielsyahputra2024kanyolo}.  They can all be effectively combined with YOLO.


\section{Conclusions}
\label{sec:conclusion}

In this article, we introduce the evolution of the YOLO series over the years, review these technologies from the perspective of modern object detection technology, and point out the key contributions they made at each stage. We analyze YOLO's influence on the field of modern computer vision from aspects such as ease of use, accuracy improvement, speed improvement, and versatility in various fields. Finally, we introduce the YOLO-related models in various fields. The purpose is that through this review article, readers can not only be inspired by the development of the YOLO series, but also better understand how to develop various real-time computer vision methods. We also hope to provide readers an idea of the different tasks YOLO can be used for and possible future directions.

\section*{Acknowledgments}
The authors wish to thank National Center for High-performance Computing (NCHC) for providing computational and storage resources.



\vfill

\end{document}